\documentclass[runningheads]{llncs}

\usepackage{eccv}

\usepackage{floatflt}

\usepackage{eccvabbrv}

\usepackage{amsmath,amsfonts,bm}

\def\eqref#1{equation~\ref{#1}}

\def\1{\bm{1}}

\DeclareMathAlphabet{\mathsfit}{\encodingdefault}{\sfdefault}{m}{sl}
\SetMathAlphabet{\mathsfit}{bold}{\encodingdefault}{\sfdefault}{bx}{n}

\DeclareMathOperator*{\argmax}{arg\,max}
\DeclareMathOperator*{\argmin}{arg\,min}

\usepackage{hyperref}
\usepackage{url}
\usepackage[utf8]{inputenc} 
\usepackage[T1]{fontenc}    
\usepackage{hyperref}       
\usepackage{url}            
\usepackage{booktabs}       
\usepackage{amsfonts}       
\usepackage{nicefrac}       
\usepackage{microtype}      
\usepackage{xcolor}         
\usepackage{enumitem}
\usepackage{graphicx}
\usepackage{dsfont}

\newcommand{\bp}{\mathbf{p}}

\newcommand{\bx}{\mathbf{x}}
\newcommand{\by}{\mathbf{y}}
\newcommand{\bz}{\mathbf{z}}

\newcommand{\btheta}{\boldsymbol{\theta}}

\newcommand{\bomega}{\boldsymbol{\omega}}

\makeatletter
\DeclareRobustCommand\onedot{\futurelet\@let@token\@onedot}
\def\@onedot{\ifx\@let@token.\else.\null\fi\xspace}

\makeatother

\definecolor{darkgreen}{rgb}{0,0.7,0}

\usepackage{algorithm}
\usepackage{algorithmic}

\usepackage{multirow}
\usepackage{amsmath}
\usepackage{xspace}
\usepackage{wrapfig}
\newcommand{\tablestyle}[2]{\setlength{\tabcolsep}{#1}\renewcommand{\arraystretch}{#2}\centering\footnotesize}
\newlength\savewidth

\def \x{\mathbf{x}}

\def \z{\mathbf{z}}

\newcommand{\methodName}{DCL\xspace}

\usepackage[accsupp]{axessibility}  

\usepackage{hyperref}

\usepackage{orcidlink}

\begin{document}

\title{Deep Companion Learning: Enhancing Generalization Through Historical Consistency} 

\titlerunning{Deep Companion Learning}

\author{Ruizhao Zhu\orcidlink{0009-0001-9496-3144} \and
Venkatesh Saligrama\orcidlink{0000-0002-0675-2268} }

\authorrunning{R Zhu and V Saligrama.}

\institute{Boston University, Boston MA, 02215, USA \\
\email{\{rzhu,srv\}@bu.edu}}


\maketitle
\begin{abstract}
\noindent
We propose Deep Companion Learning (DCL), a novel training method for Deep Neural Networks (DNNs) that enhances generalization by penalizing inconsistent model predictions compared to its historical performance. To achieve this, we train a deep-companion model (DCM), by using previous versions of the model to provide forecasts on new inputs. This companion model deciphers a meaningful latent semantic structure within the data, thereby providing targeted supervision that encourages the primary model to address the scenarios it finds most challenging. We validate our approach through both theoretical analysis and extensive experimentation, including ablation studies, on a variety of benchmark datasets (CIFAR-100, Tiny-ImageNet, ImageNet-1K) using diverse architectural models (ShuffleNetV2, ResNet, Vision Transformer, etc.), demonstrating state-of-the-art performance.
\end{abstract}    
\section{Introduction}
\begin{figure*}[t]
  \begin{center}
    \includegraphics[trim=0cm 0cm 0cm 0cm, clip,width=0.98\textwidth]{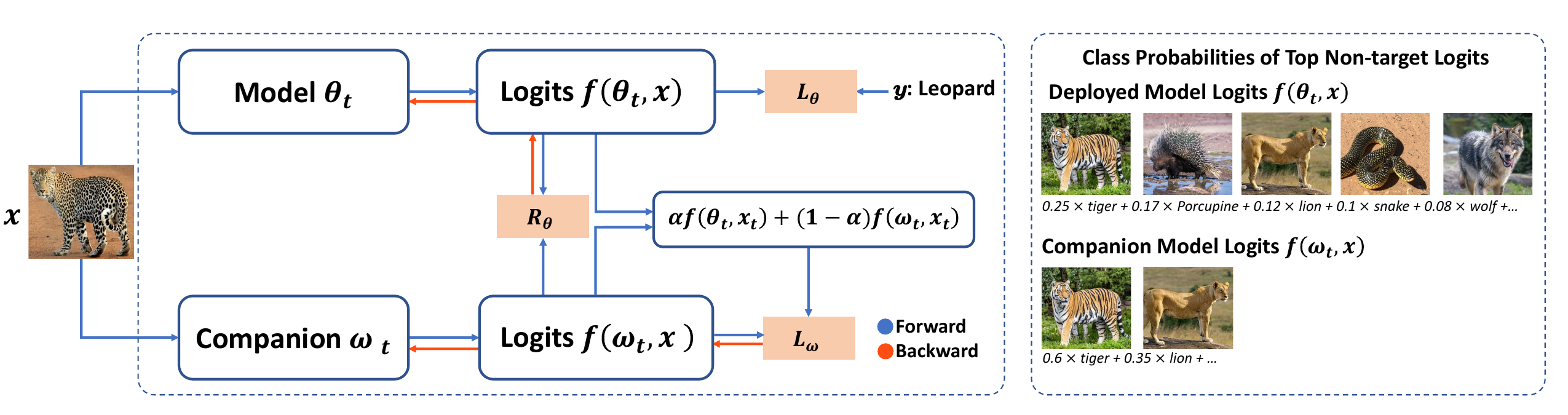}
  \end{center}

  \caption{\textbf{Method Overview.} At iteration $t$, we optimize the instantaneous model $\btheta_t$ (model eventually deployed upon training) with standard cross entropy loss and a regularizer enforcing consistency with model $\bomega$. Model $\bomega$ is recursively updated by approximating predictions from its previous embodiment and the predictions of current model $\btheta_t$. (Right) Probability of class as the top non-target class is shown. The companion model helps narrow down the top non-target classes as tiger and lion for class leopard. In the initial training stage, the top non-target of deployed model logits are more randomly distributed with some irrelevant classes. The companion model can help capture a general semantic structure of the dataset.}
  \label{fig:intro}

\end{figure*}
Stochastic Gradient Descent (SGD) underpins Deep Neural Networks (DNNs) training at scale. We propose a novel training method to improve SGD generalization in the context of supervised learning. While there are a number of prior works that have focused on improving SGD generalization, as we point out later, our perspective is shaped by the view that SGD is inherently stochastic\footnote{Due to random initialization and random choices involved in batch processing \cite{pmlr-v134-neu21a,bottou}.}, and we posit that controlling the variability of SGD trajectory during training can improve generalization. There are two sources of variability:

\begin{itemize}[leftmargin=16pt]
 \setlength\itemsep{-1.em}
\item[(a)] Due to variability across different batches, a model can observe a wide deviation in empirical losses on two different batches, and as such the observed loss on any batch may not be indicative of the true loss.\\
\item[(b)] Due to the randomness of batches, the SGD model trajectory is stochastic, and so on a new batch, the model prediction can be widely diverging depending on which SGD model was realized prior to that round. 
\end{itemize}
Our focus in this paper is on (b) as we believe the solution to (a) can benefit from more data. While data augmentation can serve as a solution for improving generalization, we view that effort as complementary to our study. To account for (b) we propose to penalize predictions made by model updates that deviate significantly from our forecast. To achieve this, we train a deep-companion model (DCM), by using previous versions of the model to provide forecasts on new inputs. In this context, our goal poses two fundamental challenges:

\begin{itemize}[leftmargin=16pt]
\setlength\itemsep{-1.em}
\item[(i)] How to best utilize past history to forecast outputs on new input examples?\\
\item[(ii)] How can we learn to make predictions efficiently?
\end{itemize}
The challenges in (i) includes: (a) designing a good look-back horizon to balance recency with historical trends, and how to use these trends to forecast; (b) what is the latent space where such predictions make most sense. The challenge in (ii) requires that our method does not significantly expand the storage or computational footprint of vanilla SGD. 

\noindent\textbf{Deep Companion Model.} 
We address (ii) through a companion neural network that aims to identify a prediction that minimizes the disagreement between itself and the preceding models. 
This companion network mirrors the architecture of the primary model (ablations with smaller networks appear in supplementary) currently undergoing training. We perform analogous SGD steps to train the companion model aligning its optimization process with that of the primary model. We use an exponential smoothing parameter and hyperparameter tune it to optimize the look-back horizon. We supervise the companion in the logit space (before softmax) by minimizing the mean-squared error. Intuitively, this makes sense because we expect well-clustered and linearly separable features in the logit space. 

\noindent \textbf{Enforcing Predictive Consistency.} Our proposal is depicted in Fig.~\ref{fig:intro}. We measure the difference between DCM output, $f(\bomega_t, \bx_t)$ and SGD model, $f(\btheta_t, \bx_t)$, and use this difference as a penalty. Our intuition here is related to the notion of cumulative regret, a concept arising in streaming settings~\cite{uml}. Although our measure is not a true measure of regret\footnote{every input sample in our case has been previously observed}, the DCM output can be viewed as a prediction on a new batch without the benefit of hindsight, while the SGD model output reflects the best achievable with hindsight of the new batch data. Intuitively, significant deviations are likely a result of SGD overfitting to the current batch, and performance can be improved by penalizing inconsistency. A different perspective is depicted in Figure \ref{fig:intro} (right). The companion model consistently outputs the same top non-target class, capturing a generalizable semantic structure of the dataset. In particular, for the class Leopard, the companion model has class Tiger or Lion as the top non-targets. 

\noindent {\bf Experimental Results.} We run experiments on several benchmark datasets (CIFAR-100, TinyImageNet, ImageNet-1K) and architectures (ShuffleNetV2, Resnet-18, Resnet-50, and ViT-Tiny) and show that our proposed input and model consistency proxies 
lead to improved SOTA performance. In particular, on CIFAR-100, our results, obtained without any pre-training, attain performance gains larger than those that utilize pre-training. In general, pre-training results in a computational bottleneck while adapting to target data. Our proposed method suggests that these bottlenecks can be overcome through a better-chosen training scheme. Our method also scales to ImageNet-1k dataset with transformer based architecture. Addtionally, DCL demonstrates its potential as a plug-and-play technique in various applications such as fine-tuning, self-supervised pre-training, and semi-supervised learning.

\noindent{\bf Contributions.} The main contributions of our paper are:

\begin{itemize}[leftmargin=8pt]
\item {\it Efficient Consistency Predictor}. We propose a computationally efficient method that uses predictions of previous model versions to forecast consistent outputs on new examples. The companion model infers a meaningful semantic structure. 
\item {\it Data-Dependent Dynamic Regularization.} Our regularizer penalizes deviations of its predictions from the companion model predictions. Since the companion model is updated in parallel, the regularization induced is dynamic, and since it penalizes predictions rather than parameters, it is data-dependent. 
\item {\it Improved Representation.} Our choice of logit space enforces better linearly separability of different classes resulting in better representations based on effectively inferring the underlying semantic structure.
\item \textit{Empirical Results.} We demonstrate SOTA performance on diverse benchmark datasets and architectures. We show that training from scratch achieves similar accuracies as models with pre-training, thereby overcoming the computational overhead of adapting pre-training to target data. 
\end{itemize}
\section{Related Work}
Various regularization techniques have been employed to enhance model generalization. These include data augmentations \cite{devries2017cutout,cubuk2018autoaugment, zhang2017mixup, hendrycks2019augmix,yun2019cutmix,muller2021trivialaugment}, dropout regularization \cite{srivastava2014dropout}, normalization \cite{ioffe2015batch, ba2016layer, wu2018group} and penalty functions \cite{loshchilov2017decoupled,krogh1991simple}.

\noindent \textbf{Parameter Regularization}
Penalty functions, in particular, are integrated with the primary loss function and jointly optimized. These functions are carefully crafted to induce specific properties to the loss function. For example, penalizing $L^2-$norm on parameters is traditionally adopted to mitigate overfitting. Furthermore, modifications to the loss landscape geometry have been proposed, with some regularizers targeting sharpness\cite{foret2021sharpnessaware,du2022sharpness,zhang2023gradient}. These regularizers capture specific local properties of the loss landscape over the parameter space, and as such are data-agnostic. We propose a method that enforces penalty on the predicted outputs, and thus inducing a data-dependent regularization. 

\noindent\textbf{Data-Dependent Regularization} 
Our method is most closely related to prior works that explicitly or implicitly induce data-dependent regularization. Variance reduction methods \cite{bottou} propose to reduce variance by using a control variate derived from a previously stored anchor model. While the idea of variance reduction is related to ours, they are evidently ineffective for deep models \cite{bottou}. In contrast, we train a companion model that is continuously updated to provide consistent forecasts, and as such is more effective. 
Similar to our approach, \cite{izmailov2018averaging} aims to reduce variance by averaging over stochastic gradient models. Other methods, such as \cite{MeanTeacher2017, BYOL2020EMA, DINOcaron2021emerging, hochreiter1997flat, foret2021sharpnessaware, cha2021swad, chen2020mocov2, he2019moco}, focus on achieving consistency across different model views. Specifically, Mean Teacher \cite{MeanTeacher2017} uses an exponential moving average (EMA) of model parameters as an anchor to interact with the current model. Unlike our method, which combines predictions, EMA fuses model parameters directly.
Temporal Ensembling \cite{laine2017temporal} uses EMA predictions for each training instance at each epoch. PS-KD \cite{kim2021self} employs a self-distillation approach, utilizing a previous model as a teacher to provide soft labels for the student model. Unlike us, they do not update the teacher model. Self-supervised learning methods \cite{DINOcaron2021emerging, BYOL2020EMA} propose similar penalties in the absence of ground truth. These works manipulate the loss landscape in parameter space like EMA \cite{izmailov2018averaging, MeanTeacher2017}, differing from our approach. DML \cite{zhang2018deep} trains two models simultaneously with different initializations, each model alternately serving as a teacher with pseudo-labels. In contrast, we start with a single initialization and learn a companion model to capture the mean behavior in the logit space, reshaping the feature representation for the deployed model to be more compactly clustered. Our regularizer adapts to both historical and recent predictions, creating a surrogate for controlling variability.
 \section{Method}

In this section, we propose Deep Companion Learning (DCL) that utilizes a regularizer to enforce prediction consistency during SGD training. 
First, we describe our method in Section \ref{subsec:dcl} and the algorithm and pseudo-code in Section \ref{subsec:implementation}. Subsequently, in Section \ref{subsec:intuition} we present an intuitive justification for our method. 
\subsection{Deep Companion Learning Method}
\label{subsec:dcl}
\textbf{Notation.} For simplicity, we will focus on a $K-$class classification problem with $\mathcal{X}$ and $\mathcal{Y}$ being the input and output spaces respectively. A training set of $N$ i.i.d. data points $\mathcal{D}=\{\mathbf{x}_i, y_i\}^{N}_{i=1}$ sampled from a joint distribution $P$ on ${\cal X} \times {\cal Y}$ is provided, where $\mathbf{x}_i \in \mathcal{X}$ and $y_i \in \mathcal{Y}$. Let $P^N$ be a distribution of any $N$-sample dataset. We parameterize the neural network with parameters $\btheta$. Given an data sample $(\bx_i, y_i)$, where $\mathbf{x}_i \in \mathcal{X},\mathbf{y}_i \in \mathcal{Y}$, $f(\btheta, \mathbf{x}_i)$ denotes the network output logits, and $\ell\left(f(\btheta, \mathbf{x}_i),y_i\right)$ denotes the loss. We consider the empirical risk minimization as $\min_{\btheta} \frac{1}{N}\sum_i \ell\left(f(\btheta, \bx_i),y_i\right) := \min_{\btheta}\mathcal{L}(\btheta)$. We construct another network with the same architecture, but parameterized differently with $\bomega$. Similarly, $f(\bomega, \bx_i)$ represents the output logits for input $\bx_i$ under the $\bomega$ model. We refer to $\bomega$ as a companion model. Subscript $t$ represents the step of the iteration.

To build intuition into our method, let us consider a data sample as a triple, $(\bx_i,y_i,\bz_i)$ consisting of the input $\bx_i$, the ground-truth label $y_i$, and an auxiliary observation $\bz_i$, for instance, logits obtained from an auxiliary network on the input $\bx_i$. What would be our goal in this case? Naturally, we would like to train our model $\btheta$ by including the auxiliary observation as supervision. To this end, we define a new loss, $\Delta(f(\btheta, \bx_i),\bz_i)$, which denotes the distance between logits predicted by $\btheta$ and the auxiliary observation. As such our global objective $R(\btheta)$ is to optimize:
\begin{align}
\label{eq:theta_update}
R(\btheta) = \frac{1}{N}\sum_{i=1}^N r(\btheta,\bx_i,y_i,\bz_i) \triangleq 
\frac{1}{N}\sum_{i=1}^N [\ell(f(\btheta, \bx_i), y_i) + \lambda \Delta (f(\btheta, \bx_i),\bz_i) ]
\end{align}
where $\lambda$ is a hyperparameter.

Let us now describe SGD in this context. In round $t$ nature chooses an example $(\bx_t,y_t,\bz_t)$ uniformly at random from the dataset, and a corresponding risk function $r_t(\btheta) = r(\btheta,\x_t,y_t,\z_t)$. SGD then takes a gradient step on the observed risk, namely,
\begin{align}
    \label{eq:theta_update_new}
    \btheta_{t+1}= \btheta_t - \eta \nabla_{\btheta} (\ell(f(\btheta_t, \bx_i), y_i) + \lambda \Delta(f(\btheta_t,\x_t), \z_t)
\end{align}
\subsection{Companion Model}
Let us now discuss training a companion model, $\bomega$ to predict auxiliary logits, $\z=f(\bomega,\x)$, which serve as supervision for our deployed model update above.At round $t$, we have historical information upto round $t-1$, and the goal of an auxiliary model is to offer a forecast for the logits corresponding to the new input, $\x_t$ in round $t$. As such it makes sense for the companion model to provide supervision that complements the ground truth information $y_t$. A direct choice is to encourage the companion model to be close to all the historical models  $\{\btheta_t, \btheta_{t-1}, ..., \btheta_1\}$, we design the companion objective as in Equation \ref{eq:omega_update}. 
\begin{align}\label{eq:omega_update}
    \bomega_t = \argmin_{\bomega} \left(\sum_{i=1}^{t}\Delta(f(\bomega, \bx_t), f(\btheta_i, \bx_t))\right)
\end{align}

The update rule in~\ref{eq:omega_update} requires all historical models, which means that the memory complexity grows linearly as training iterates, and thus impractical. When the distance function is MSE, i.e. $\Delta(f(\btheta, \bx), f(\bomega_t, \bx)) = \frac{1}{2}\|f(\btheta, \bx) - f(\bomega_t, \bx)\|^2$, the well-known orthogonality property leads to the following observation:
\begin{align*}
\frac{1}{t}\sum_{i=1}^t \|f(\btheta_i, \bx) - f(\bomega, \bx)\|^2 =& \|f(\bomega, \bx) - \frac{1}{t} \sum_{i=1}^t  f(\btheta_i, \bx)\|^2 + \text{Var}(f(\btheta_i, \bx))
\end{align*}
The term $\frac{1}{t} \sum_{i=1}^t  f(\btheta_i, \bx)$ can be expressed recursively assuming the companion model in the previous round is a good approximation:
\begin{align*}
\frac{1}{t} \sum_{i=1}^t  f(\btheta_i, \bx) = \frac{t-1}{t}  f(\bomega_{t-1}, \bx) + \frac{1}{t}  f(\btheta_t, \bx) + \text{noise}
\end{align*}
Ignoring the noise term, and substituting hyperparameters $\alpha$ in place of $\frac{t-1}{t},\frac{1}{t}$ we can replace Equation \ref{eq:omega_update} with the following objective: 
\begin{align}
\label{eq:omega_update_new}
 \bomega_{t} = \argmin_{\bomega} \frac{1}{2}\| f(\bomega, \bx), \alpha f(\bomega_{t-1}, \bx) + (1-\alpha) f(\btheta_{t}, \bx) \|^2
\end{align}

The objective then reduces to aligning the logits of the current companion model with a convex combination of the output of the previous companion model, and the output of the instantaneous deployed model.

\subsection{Implementation}
\label{subsec:implementation}
\noindent 

\noindent \textbf{Algorithm.} The end-to-end pseudo code is displayed in Algorithm~\ref{alg:constrained_learning}. Deep Companion Learning (DCL) algorithm, iteratively trains the (deployed) model as in Equation \ref{eq:theta_update_new} and the companion model as in Equation \ref{eq:omega_update_new}, by taking a gradient step on the batch data.

\begin{algorithm}[H]
   \centering
   \caption{Deep Companion Learning (\methodName)}
   \label{alg:constrained_learning}
\begin{algorithmic}[1]
   \STATE \textbf{Input:} Training data $\mathcal{D} = \{ (\mathbf{x}_{i}, y_{i}) \}^{N}_{i=1}$
   \STATE \textbf{Parameters:} Iteration $T$,  batch size $B$, learning rates $\eta_\theta$, $\eta_\omega$, companion weight $\alpha$
   \STATE \textbf{Initialize:} Randomly initialize model $\btheta_0$, $\bomega_0$ with the same initialized parameters.
   \FOR{$t=0$ \textbf{to} $T-1$}
   \STATE {Sample a batch of data $\{\bx, \by\}$}
   \STATE  Update the instantaneous deployed model:\\
       $\btheta_{t+1}= \btheta_t - \eta_{\btheta} \nabla_{\btheta} (\mathcal{L}(\btheta) + \Delta(f(\btheta,\bx), f(\bomega_{t},\bx)))$
   \STATE Update companion model:\\
$\bomega_{t+1} = \bomega_{t} - \eta_{\bomega} \nabla_{\bomega} \left [\Delta(f(\bomega), \alpha f(\bomega_{t}) + (1-\alpha) f(\btheta_t)  ) \right]$
   \ENDFOR
   \STATE \textbf{Return :} $\btheta_T$
\end{algorithmic}
\end{algorithm}

\noindent{\bf Other Learning Settings}. In addition to supervised learning, \methodName can also be applied to other settings. For instance, we can employ \methodName to fine tune a pre-trained model on different downstream tasks. This is possible because the companion model does not use ground-truth labels for training, and thus can leverage unlabelled datasets. This naturally leads to a semi-supervised learning setting. DCL can also be employed for self-supervised pre-training. For example, we can employ DCL for Masked Autoencoder (MAE) \cite{he2022masked} training. Here in lieu of cross-entropy loss used in classification, we replace the loss $\ell$ in Equation \ref{eq:theta_update} to reconstruction loss. Similarly for the regularizer, we enforce consistency for the reconstructed output images. DCL can also be employed for Knowledge Distillation (KD) by adding an additional regularizer to the student during training. Intutitively, the student then seeks a consensus with both the teacher and the companion model.

\subsection{Intuitive Justification}
\label{subsec:intuition}
\begin{figure}[t]
  \begin{center}
    \includegraphics[trim=0cm 0.0cm 0cm 0cm, clip,width=1\textwidth]{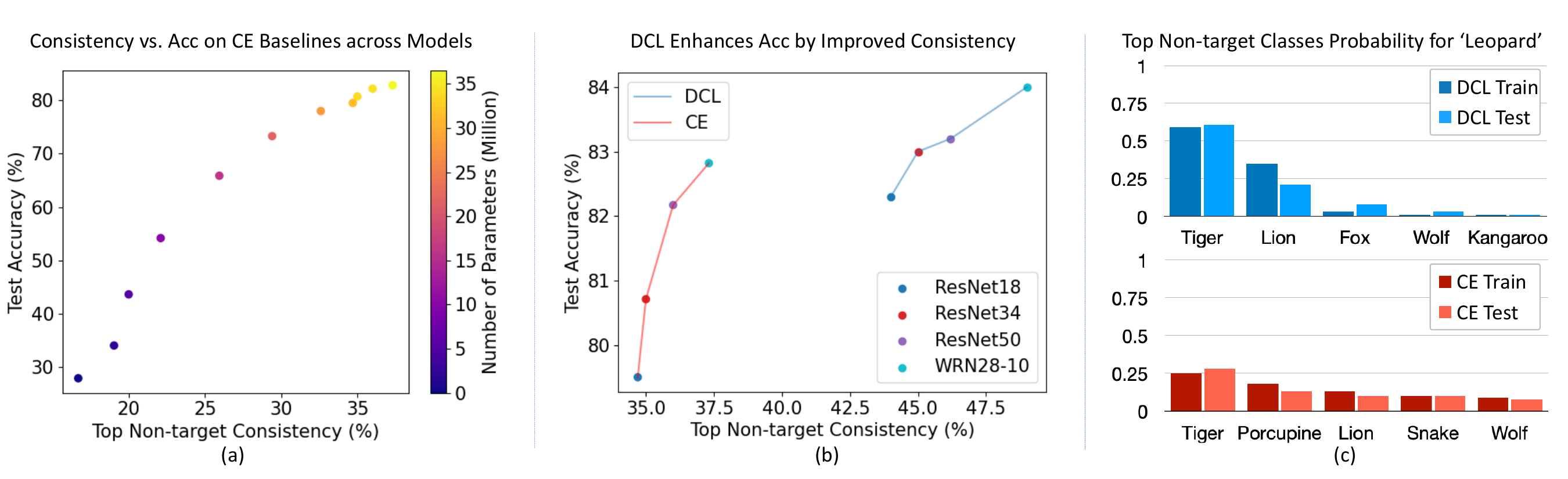}
  \end{center}
  \caption{ \textbf{Higher Top Non-target Consistency Indicates Better Generalization.} We visualize training top non-target consistency and test accuracy with CIFAR-100 across different models. (a) Larger CE models have better generalization while having larger top non-target consistency. This indicates a positive correlation between top non-target consistency and test accuracy. (b) Across different architectures we see a consistent correlation of improved DCL test accuracy with increasing top non-target consistency. (c) DCL chooses Tiger and Lion most frequently as the top non-target classes of Leopard during training while CE exhibits inconsistent patterns.
}
  \label{fig:Consistency}
\end{figure}

\noindent{\bf Consistency of top non-targets in training is correlated with improved generalization.} We highlight salient aspects of DCL using experiments on CIFAR-100 across various architectures in Figure \ref{fig:Consistency}. The training data consists of $N$ samples of $K$ classes with $N_c$ data samples for each class $c$. For data $(\bx, y)$, we define the predicted logits from the model $\btheta$ as $f(\btheta,\bx) \in \mathbb{R}^{K}$, where the $k$-th digit $f(\btheta,\bx)_k$ is the predicted probability of the $k$-th class. Then we define the top non-target class for the specific data as $\Bar{c}_{\bx} = \argmax_{k, k\neq\by} f(\btheta,\bx)_k$, namely the class with the largest logits value among all the non-target classes. For a specific class $c$, its top non-target class consistency is defined as $\max_{k,k\neq c} \frac{\sum_{\bx, y=c}\mathds{1} (\Bar{c}_{\bx} = k)}{N_c}$, the percentage of the most frequent occurring top non-target class. For instance, Figure \ref{fig:Consistency} (c) demonstrates for the class Leopard, the top non-target class is the Tiger with around $60\%$ probability while the class Lion is $30\%$ for both training and testing. This metric reflects how consistent the top non-target class is among the data of the same class. DCL effectively narrows the choices among the top non-target classes to tiger and lion while CE baseline evidently fails to capture this fine-grained semantic structure, with the probability of tiger or lion bearing similarity to other less meaningful classes. Therefore, the training consistency reveals the underlying semantic structure of the dataset in the logits space, which generalizes to test data as well. As expected, Figure \ref{fig:Consistency} (a) depicts a positive correlation between training consistency and testing accuracy and  Figure \ref{fig:Consistency} (b) shows DCL can improve training consistency leading to improved test accuracy across diverse architectures. 

\begin{figure*}[t]
  \begin{center}
    \includegraphics[trim=0cm 0cm 0.2cm 0cm, clip,width=1\textwidth]{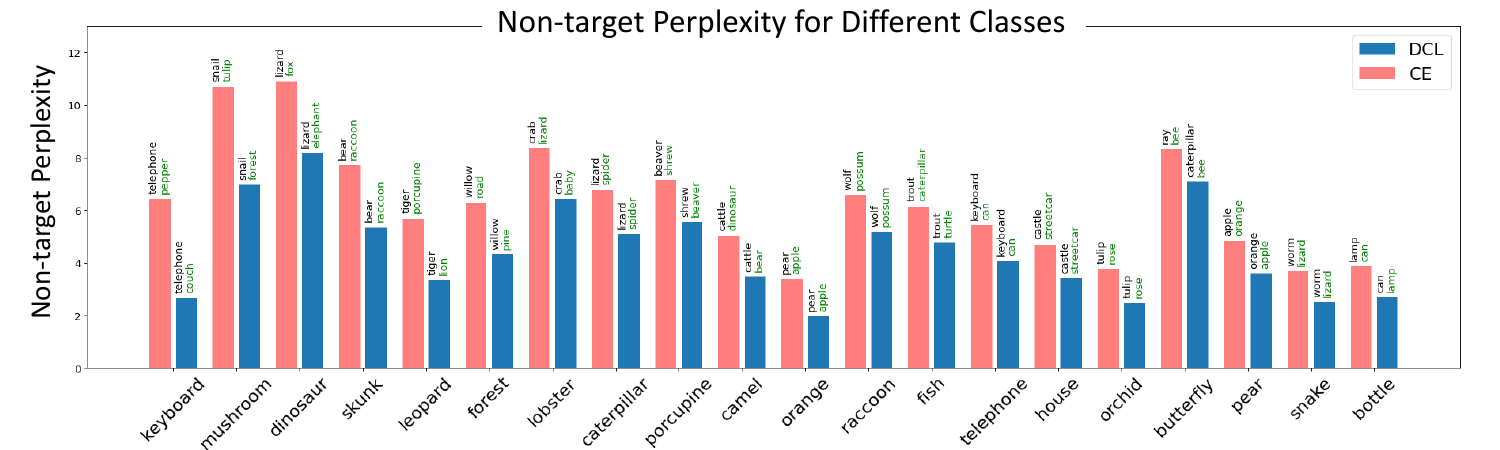}
  \end{center}
  \caption{\textbf{Non-target Perplexity for Different Classes.} The most (black text) and second most (green text) frequent classes as the top non-target class are shown on each bar. DCL can reduce perplexity over CE baseline.}
  \label{fig:perplexity_bar}
\end{figure*}

\noindent{\bf Perplexity of non-targets class is lower for DCL compared with CE.}
In addition to examining the top non-target class, we also analyze the behavior of all non-target classes. We introduce non-target perplexity for each class, which is the perplexity over the conditional distribution of non-target classes occurring as the top non-target class, given a target class $c$. We define this conditional distribution $p(k|c) = \frac{\sum_{\bx, y=c}\mathds{1} (\Bar{c}_{\bx} = k)}{N_c}, \forall k\in\{1,..,K \}, k\neq c$. 
We then define the perplexity $PP_c=\prod_{\forall k\in\{1,..,K \}, k\neq c}p(k|c)^{-p(k|c)}$. Lower perplexity indicates more consistency. 
Figure \ref{fig:perplexity_bar} compares the non-target perplexity between DCL and CE for 20 randomly selected classes in ResNet18 CIFAR100 experiments. Considering the class 'forest' as an example, the two most frequent top non-target classes identified by CE are 'willow' and 'road', whereas for DCL, they are 'willow' and 'pine'.  'Pine' is semantically more reasonable than 'road'. As expected, DCL can reduce perplexity consistently over all classes.

\section{Experiment}
\label{sec:experiment}
In this section, we evaluate the proposed DCL method on various datasets and architectures for different settings, and perform ablative studies.

\subsection{Experimental Setup}
\textbf{Datasets.}We consider publicly available image classification datasets: (a) CIFAR-100 \cite{Krizhevsky09learningmultiple} consists of $50$K training and $10$K test images from $100$ classes with size $32\times32\times3$, (b) Tiny-Imagenet \cite{le2015tinyImagenet} contains $100$K training and $10$K test images from $200$ classes with size $64\times64\times3$, and (c) ImageNet-1K \cite{russakovsky2015imagenet} consists of $~1.2$M training and $100$K test images from $1000$ classes with size $224\times224\times3$. (d) For fine-tunning and and pre-trained downstream task, we also utilize CUB \cite{CUB}, Oxford-Pet \cite{Pets}, Food-101 \cite{Food-101} and Stanford Car \cite{cars} datasets. Same image sizes are applied for the fine-tunning datasets with the pre-trained dataset.\\
\textbf{Settings.} We apply \methodName to supervised training on classification from scratch. Later on in the application section, we also utilize \methodName for fine-tunning, semi-supervised learning, self-supervised pre-training and knowledge distillation tasks.

\noindent\textbf{Baselines.} We mainly compare \methodName with standard cross-entropy baseline (CE). We also compare \methodName with recent works with different optimization and regularization techniques \cite{foret2021sharpnessaware,du2022sharpness,zhang2023gradient}. Additionally, related baselines for self-distillation are also included \cite{zhang2018deep,kim2021self}.

\noindent\textbf{Models.} We evaluate ResNet \cite{ResnetOriginalPaper}, ShuffleNetV2 \cite{Ma_2018_ECCV_ShuffleNetV2}, ViT \cite{dosovitskiy2020image} and Swin Transformer \cite{liu2021swin} architectures on these datasets. In particular, we benchmark ResNet18, ResNet34, ResNet50, and ShuffleNetV2 models on CIFAR-100 and Tiny-Imagenet dataset. Due to computing limitations, we only train ResNet18, ResNet50, Vit-t, Swin-t on ImageNet-1k dataset. For fine-tunning dataset, we benchmark EfficientNet \cite{tan2019efficientnet} pre-trained on ImageNet. We provide their architectural details in supplementary. 
\begin{table*}[t]
\tablestyle{0.7pt}{1.2}
\caption{\textbf{Performance Comparison on CIFAR-100, Tiny-ImageNet and ImageNet-1k.} We benchmark \methodName against CE and pre-trained baselines with various models. We report Gain as accuracy difference between \methodName and CE. It clearly shows that \methodName significantly outperforms CE methods. In addition, it reaches accuracy of ImageNet pre-trained(PT) baseline without any additional data and requires far less computation. (Standard errors for experiments on CIFAR-100 are within $\pm0.26$, on ImageNet variants are within $\pm0.2$).  }
\label{table.all_model_stats_cifar_tiny_imagenet}
\centering
\begin{tabular}{l | c c c c | l| c c c | l| c c c}
\toprule
\multicolumn{5}{c|}{\textbf{CIFAR-100}}& \multicolumn{4}{c|}{\textbf{Tiny-ImageNet}} & \multicolumn{4}{c}{\textbf{ImageNet-1K}} \\
\hline
\textbf{ Model} &\textbf{ CE } &\textbf{ PT } & \bf{Ours} & \bf{Gain} & \textbf{ Model} &\textbf{ CE } & \bf{Ours} & \bf{Gain} & \textbf{ Model} &\textbf{ CE } & \bf{Ours} & \bf{Gain} \\
\hline
{ResNet18}   & {$78.3$}  & 80.5 & 82.4   &  \textbf{4.1}& {ResNet18} & {$61.8$} & 64.6  &  \textbf{2.8} & {ResNet18} &  69.5  & 70.5 & \textbf{1.0} \\ 
{ResNet34}   & {$80.6$} & 82.6 & 83.1  &  \textbf{2.5} & {ResNet34} & {$64.0$} & 66.8  &  \textbf{2.8} & {ResNet50} &  76.0   &  77.1 & \textbf{1.1 } \\
{ResNet50}   & {$82.1$} &{$82.9$}   & 83.6   &  \textbf{1.5}& {ResNet50} & {$64.4$} & 67.0 &  \textbf{2.6 } & {ViT-T} &  65.7  & 66.5 & \textbf{0.8} \\
{ShuffeV2}   & {$72.8$}  & 73.5 & 74.0   &  \textbf{1.2}& {ShuffeV2} & {$53.9$} & 55.5  &  \textbf{1.6} & {Swin-T} &  81.3  & 81.7 & \textbf{0.4} \\

\bottomrule
\end{tabular}
\end{table*}

\noindent\textbf{Hyper-parameters.} For the CIFAR-100 and Tiny-Imagenet datasets, we use the SGD optimizer with a momentum of $0.9$ and weight decay of $5e-4$. We train these models up to $200$ epochs with cosine learning rate decay with $0.1$ as the initial learning rate and batch size of $128$. For ImageNet-1k experiments, due to hardware limitations, we follow \cite{zhang2023gradient} setting using batch size of $256$, $0.1$ as the initial learning rate with cosine decay. We train ResNet for $90$ epochs and ViTs for 300 epochs. For ResNet, we use an SGD optimizer with $0.9$ as momentum and for the ViTs experiments, we use AdamW optimizer with $\beta_1=0.9$, $\beta_1=0.999$. All experiments use base augmentation on data. We provide the remaining hyper-parameter and experiments with different augmentations in supplementary.

\subsection{Results}
Table~\ref{table.all_model_stats_cifar_tiny_imagenet} compares the performance of our method and standard baseline on CIFAR-100, Tiny-Imagenet and ImageNet-1k datasets. Table~\ref{table:previous_methods} compares the performance of \methodName with recent baselines for different optimization techniques including self-distillation.  We report the accuracy of the final iteration in all the methods. Below, we highlight the main takeaway points.

\noindent \textbf{\methodName Achieves Better Generalization.} We note \methodName consistent improves upon CE across different datasets and architectures in Table \ref{table.all_model_stats_cifar_tiny_imagenet} compared to baselines. For instance, on CIFAR-100 with ResNet18 architecture, CE achieves $78.3\%$ accuracy while \methodName achieves $82.4\%$ accuracy. Experiments on Tiny-ImageNet shows a similar pattern. Table \ref{table:previous_methods} shows \methodName uniformly outperforms recent state-of-the-art methods across different backbones for the batch size and basic augmentation in \cite{8932404GAM-RHN} (Supplementary reports other augmentations).

\noindent \textbf{Scalability to Large ImageNet-1k dataset.} Table~\ref{table.all_model_stats_cifar_tiny_imagenet} shows that \methodName scales well to large datasets such as Imagenet-1K. In particular, it achieves better accuracy than the baseline. For instance, with the ResNet50 architecture, the CE method achieves $76.0\%$ accuracy while DCL achieves $77.1\%$ accuracy. \looseness=-1

\noindent \textbf{Scalability to Transformer based architecture.} Table~\ref{table.all_model_stats_cifar_tiny_imagenet} shows that \methodName scales well to different tranformer-based backbones such as ViT and Swin transformer. For instance, with the ViT-T architecture, the CE method achieves $65.7\%$ accuracy while DCL achieves $66.5\%$ accuracy. \looseness=-1

\noindent \textbf{DCL Trained-from-scratch is Superior to ImageNet Pre-Trained Models.} 
 Table~\ref{table.all_model_stats_cifar_tiny_imagenet} shows the performance of the different models pre-trained on ImageNet and fine-tuned on the CIFAR-100 dataset. It takes CIFAR-100 $32\times 32\times 3$ image and scales to $224\times224\times 3$ image and runs the inference using this input. In contrast, DCL trained the model with \methodName using only CIFAR-100 data with $32\times32\times3$ input. \methodName achieves better performance than the pre-trained counterpart. This is important because pre-training is expensive. For example when using a ResNet50 backbone, during inference \methodName trained model requires $1298$M MACs which is much lower than the $4198$M MACs required by the ImageNet pre-trained model. In addition, the proposed method requires less data to achieve this performance, i.e., only $100$K CIFAR-100 images compared to $1.2$M ImageNet images. Thus, \methodName yields faster inference and requires less sample complexity to achieve competitive performance as ImageNet pre-trained model. 

\noindent \textbf{Small \methodName Models Outperform Large CE models.} \methodName trained on small models compares favorably with CE-trained large models. For instance, on CIFAR-100, ResNet18 trained with \methodName achieves better accuracy than the much larger ResNet34 model trained with the cross-entropy method. Similar trend is evident in the context of ResNet34  vs. 
ResNet-50 performance. Furthermore, \methodName trained ResNet34 gets better performance than the larger pre-trained ResNet50 model, showing further benefits of \methodName.

\begin{table}[t]
\tablestyle{8pt}{1.2}
\caption{\textbf{Recent Baselines.} \methodName outperforms other baselines on various backones and datasets. Results reported are for the setup of GAM\cite{zhang2023gradient}, which uses basic augmentations and 256 batch size for all ImageNet experiments. SAF\cite{du2022sharpness} results on ImageNet use a batch size of 4096, an important factor for improved performance in ImageNet. Not reported here are more sophisticated augmentations, such as TrivialAugment(TA)\cite{muller2021trivialaugment}. In supplementary we show TA with CE gets $84.3\%$, consistent with that reported in \cite{muller2021trivialaugment}'s while TA with DCL $85.7\%$ on CIFAR100 for WRN28-10. For ImageNet \cite{muller2021trivialaugment} uses a batch size of 2048, significantly larger than ours.} 
\label{table:previous_methods}
\centering

\begin{tabular}{c|c|c|c}
\toprule

\multirow{2}{*}{\textbf{Methods}}& \multicolumn{2}{|c|}{\textbf{CIFAR-100}} & \bf{ImageNet-1k} \\

& \textbf{\begin{tabular}[c]{@{}c@{}}  ResNet18  \end{tabular}}   & \textbf{\begin{tabular}[c]{@{}c@{}} 
  WRN28-10   \end{tabular}} & \bf{ResNet50}
 \\ \hline\hline 
 CE\cite{zhang2023gradient}     & $78.3\pm 0.32$ & $81.4\pm 0.13$ &  $76.0\pm0.19$ \\ 
 DML  \cite{zhang2018deep}    & $79.9\pm 0.32$  & $82.7\pm0.14$ & $75.8\pm 0.15$ \\ 
 SAM  \cite{du2022sharpness}  & $79.3\pm 0.25$  & $83.4\pm0.06$ & $76.5\pm 0.11$\\ 
 PSKD \cite{kim2021self}   & $80.6\pm 0.26$  & $81.9\pm0.10$ &$76.3\pm 0.15$ \\ 
SAF  \cite{du2022sharpness}  & $80.8\pm 0.08$  & $83.8\pm0.04$ & $76.4\pm 0.15$\\ 
 GAM  \cite{zhang2023gradient}    & $80.5\pm 0.24$  & $83.5\pm0.09$ & $76.6\pm 0.19$\\ 
 Ours (MSE)   & $\mathbf{82.4\pm 0.26}$   & $\mathbf{84.2\pm0.10}$ & $\mathbf{77.1\pm 0.15}$\\ 
\bottomrule
\end{tabular}
\end{table}

\subsection{Extended Applications}

We extend \methodName to different settings. We provide several examples to show the effectiveness of our method.

\noindent {\bf Fine-tuning.} Table \ref{table:fine-tunning} shows the generalization of models when trained on sufficient labeled data and finetuned on a small dataset. We use ImageNet-1k pretrained EfficientNet as initial model and fine-tune on different smaller datasets. We outperform standard cross-entropy loss baselines. For example \methodName achieves $91.7\%$ accuracy compared with the baseline $90.9\%$ on Oxford-Pet dataset. 
\begin{table}[t]
\caption{\textbf{Fine-tuning.} Performance comparison with fine-tuning ImageNet pretrained EfficientNet. \methodName outperforms cross-entropy on different downstream tasks.} 
\label{table:fine-tunning}
\centering
\begin{tabular}{c|c|c|c|c}
\toprule
 \textbf{\begin{tabular}[c]{@{}c@{}}Fine-tunning Methods\end{tabular}}  & 
 \textbf{\begin{tabular}[c]{@{}c@{}}\ \ Food-101\ \  \end{tabular}}  &
 \textbf{\begin{tabular}[c]{@{}c@{}}\ \  CUB-200\ \  \end{tabular}}  &
 \textbf{\begin{tabular}[c]{@{}c@{}} Oxford Pet \end{tabular}}  &
 \textbf{\begin{tabular}[c]{@{}c@{}} Stanford Car \end{tabular}}  
 \\ \hline\hline 
   CE  & 82.5 & 63.8& 90.9 & 78.2 \\ 
     \methodName  & \bf 85.0 & \bf 64.7 & \bf 91.7 & \bf 79.0\\ 
\bottomrule
\end{tabular}
\end{table}

\noindent {\bf Semi-supervised Learning.} 
\methodName can be applied on top of existing semi-supervised learning. While several methods, such as FixMatch \cite{FixMatchConsistency}, propose a consistency concept based on various input augmentations, \methodName is focused on achieving consistency across different models realized along the training trajectory. 
This unique perspective allows \methodName to be seamlessly integrated with existing semi-supervised learning techniques and leads to performance improvement. 
\begin{wrapfigure}[14]{r}{0.5\textwidth}
  \begin{center}
    \includegraphics[trim=0.2cm 0.2cm 0cm 0cm, clip,width=0.45\textwidth]{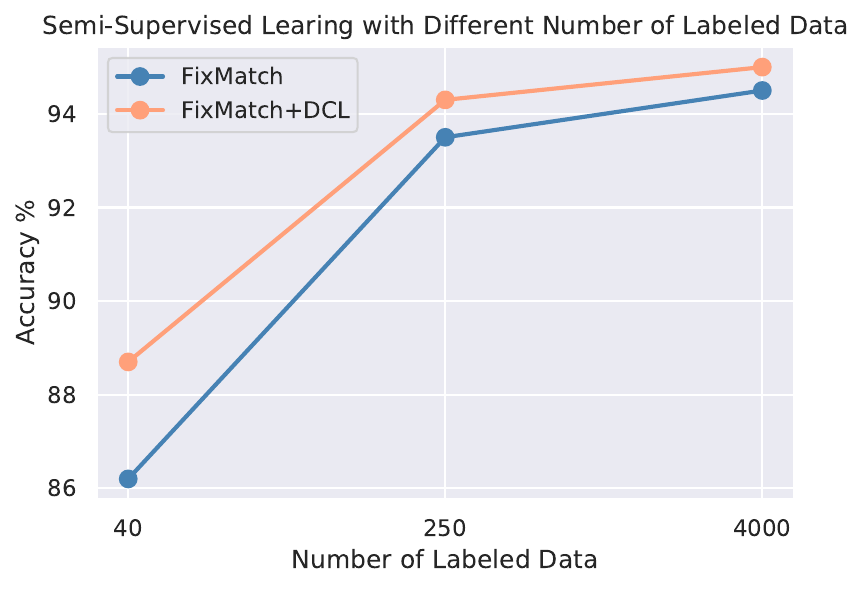}
  \end{center}
    \caption{\textbf{Semi-Supervised Learning.} \methodName demonstrates superior performance over FixMatch, particularly with fewer labels.}
  \label{fig:semi}
\end{wrapfigure}
We simply add our regularizer to their loss function and update $\btheta$. This is possible because the companion model does not require labels. Figure \ref{fig:semi} shows comparison of FixMatch with and without \methodName. With \methodName, FixMatch gains $1\%$ to $3\%$ accuracy on CIFAR-10 classification with different number of labels. In particular, scenarios we gain more for fewer labels. For example, with only 40 labeled data, \methodName can improve performance of FixMatch from $86.2\%$ to $88.7\%$. All experiments are conducted with WRN-28-2 backbone with the same hyperparameters.

\noindent {\bf Self-Supervised Pretraining.} Instead of standard cross-entropy loss for classification task, self-supervised learning often focuses on different pretext tasks for pre-training. For example, the Masked Autoencoder (MAE) \cite{he2022masked} is a variant of self-supervised learning that learns to predict or reconstruct the original images from partially masked or corrupted images. We replace the loss $\ell$ in Equation \ref{eq:theta_update} with the reconstruction loss. For the regularizer, we enforce consistency with respect to reconstructed images. Due to computational limitation, we only finetune the pretrained model on classification tasks on several small datasets. Table \ref{table:pretraining} shows utilizing \methodName to train MAE leads to better representations for downstream tasks. For example, \methodName achieves $80\%$ accuracy on CUB but vanilla MAE gets $79.2\%$. All experiments are conducted using ViT-Base backbone and we provide details for the hyperparameters in the supplementary.
\begin{table}[h]
\caption{ \textbf{Self-Supervised Pretraining.} Performance of MAE \cite{he2022masked} on ViT-B backbone with and without \methodName with fine-tunning downstream classification tasks. We use ImageNet-1k for pre-training and we use Tiny-ImageNet and other smaller dataset for fine-tunning. MAE with \methodName outperforms plain MAE, showing benefits of \methodName extension on self-supervised pre-training framework.} 
\label{table:pretraining}
\centering
\begin{tabular}{c|c|c|c|c|c}
\toprule
 \textbf{\begin{tabular}[c]{@{}c@{}}Methods\end{tabular}}  & 
 \textbf{\begin{tabular}[c]{@{}c@{}} ImageNet \end{tabular}}  &
 \textbf{\begin{tabular}[c]{@{}c@{}} Food-101 \end{tabular}}  &
 \textbf{\begin{tabular}[c]{@{}c@{}} CUB-200 \end{tabular}}  &
 \textbf{\begin{tabular}[c]{@{}c@{}} Oxford Pet \end{tabular}}  &
 \textbf{\begin{tabular}[c]{@{}c@{}} Stanford Car \end{tabular}}  
 \\ \hline\hline 
MAE \cite{he2022masked}  & 82.8 & 87.8 & 79.2 &91.5 & 82.5  \\ 
MAE+DCL & \bf 83.4 & \bf 88.8 & \bf 80.0 &\bf 92.0 & \bf 87.2 \\ 
\bottomrule
\end{tabular}
\end{table}

\noindent \textbf{Knowledge Distillation (KD).} Knowledge Distillation similarly employs the concept of aligning the output distribution of two models, but alignment is achieved by the interaction between a smaller student network and a larger pre-trained teacher network. We apply DCL in addition to different KD methods in Table \ref{table:kd}. \methodName with student get $74.5\%$ better than basic KD\cite{hinton2015distilling} without even employing the large teacher model. \methodName can also assist other SOTA methods\cite{chen2021reviewkd,chen2022simkd,zhao2022decoupled} achieving enhanced performance.
\begin{table}[h]
\caption{\textbf{Knowledge Distillation}. We shows results of ResNet8$\times$4 as student and trained ResNet32$\times$4 teacher model. \methodName without trained teacher network outperforms KD\cite{hinton2015distilling}. DCL can improve other SOTA knowledge distillation methods \cite{chen2021reviewkd,chen2022simkd,zhao2022decoupled}.} 
\label{table:kd}
\centering
\begin{tabular}{c|cc|cccc}
\toprule
&\bf{Teacher} & \bf{Student}  & \bf{\quad KD}\cite{hinton2015distilling}\quad\quad & \bf{ReviewKD}\cite{chen2021reviewkd} & \bf{SimKD}\cite{chen2022simkd} & \bf{DKD}\cite{zhao2022decoupled} \\
\hline\hline 
w/o DCL& 79.4 & 72.5 & 74.0 & 75.6 & 77.8 & 75.9 \\
w/ DCL& - & \textbf{74.5} & \textbf{75.2}  & \textbf{76.2}  & \textbf{78.0}  & \textbf{76.5} \\
 \bottomrule
\end{tabular}
\end{table}

\begin{figure*}[t]
    \centering 
    \hfill
        \centering
        \includegraphics[trim=0cm 0.2cm 0.5cm 0.5cm,width=1.05\textwidth]{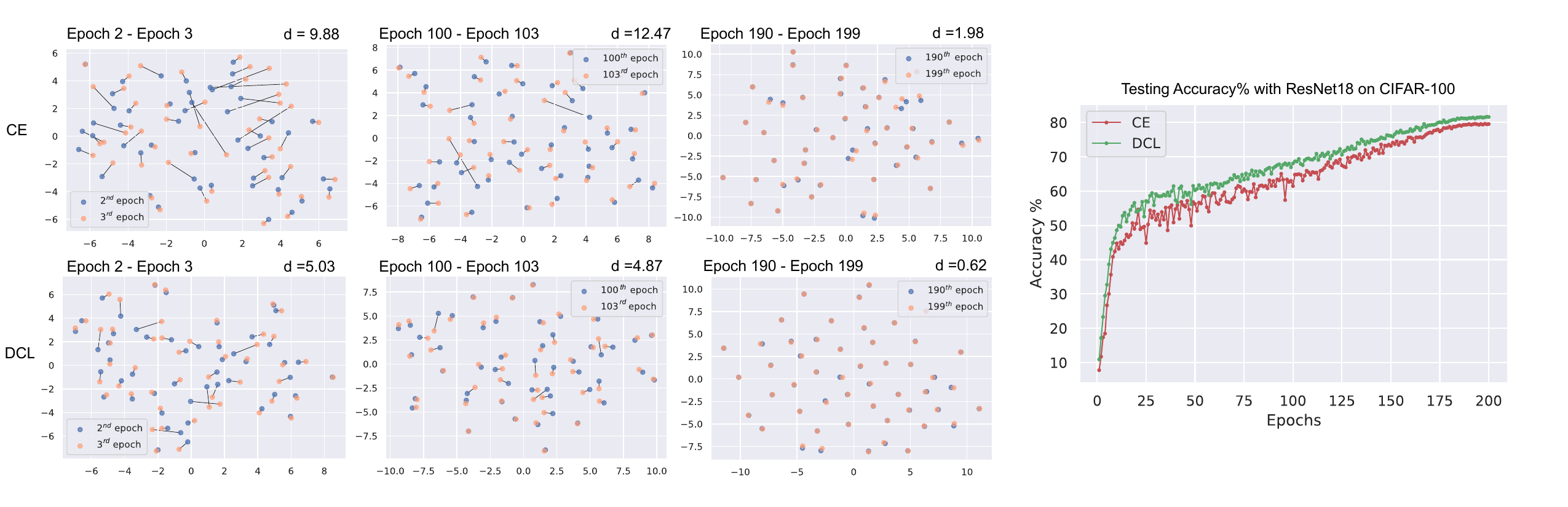}
    \caption{\textbf{Comparison of Model Variation and Test Accuracy along Training Trajectory on CIFAR-100 with ResNet18.} The left section displays t-SNE visualizations of output logits for 50 test data samples, illustrating reduced variation in logits output across various training stages using \methodName. In the plot, $d$ is the average distance over data representation in the logit space. The right section presents the progression of test accuracy during training, DCL predictions show smaller variation and attain better generalization. } 
    \label{fig:variance}
\end{figure*}

\noindent{\bf \methodName Reduces Model Variation and Improves Generalization.} Figure \ref{fig:variance} shows t-SNE visualization of logit space for test data to compare model variation and test accuracy along training ResNet18 on CIFAR-100 dataset. The figures shows changes of each data logits between neighboring epochs. At different stages of training, \methodName consistently demonstrates reduced fluctuation in the logit space. This pattern is mirrored in a smoother accuracy trajectory during training, with \methodName exhibiting superior performance compared to the CE baseline.

\subsection{Ablations}\label{subsec:ablation}

\begin{figure}[t]
  \begin{center}
    \includegraphics[trim=0.2cm 0.2cm 0cm 0cm, clip,width=0.8\textwidth]{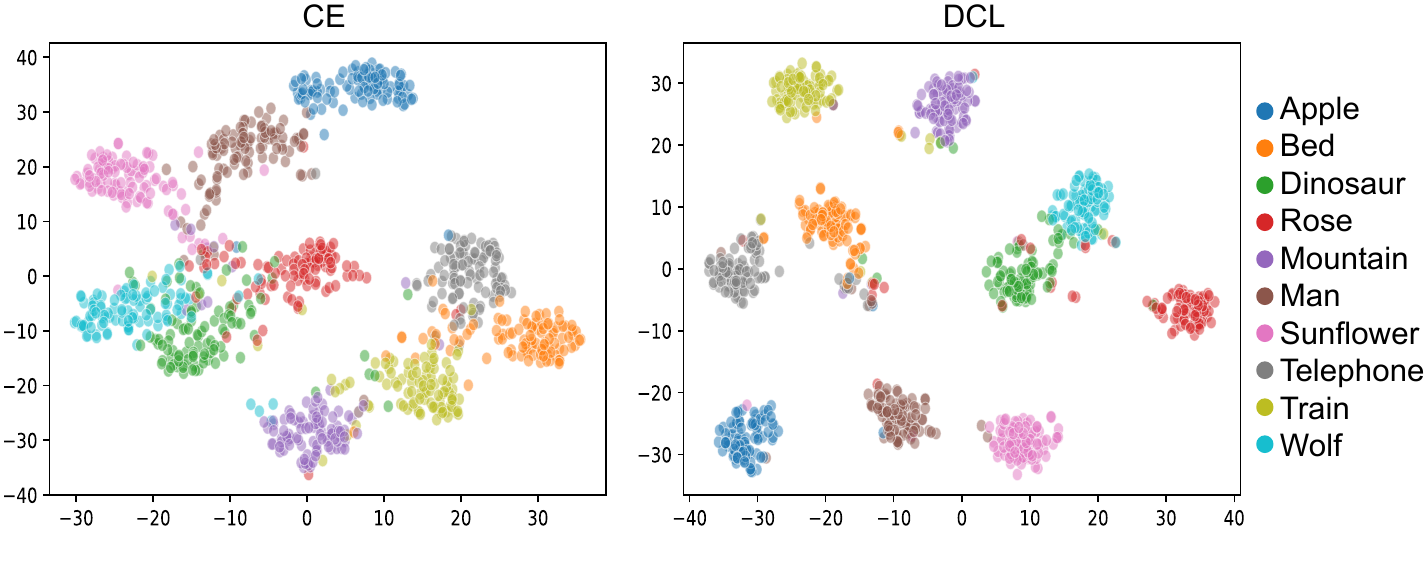}
  \end{center}
  \caption{\textbf{Logits Visualization.} t-SNE visualization of logit space from trained ResNet18 on CIFAR-100 using test data from 10 random classes.}
  \label{fig:clustering}
\end{figure}

\noindent{\bf \methodName Generates Better Logit Space Representation.} Figure \ref{fig:clustering} shows t-SNE visualization of logit space for 10 random classes of CIFAR-100 test data. \methodName induced logit space enforces better linear separability of different classes.

\noindent{\bf Different Distance Functions.} Instead of Mean Square Error as the distance function $\Delta$, we can also use other forms of distance. Table \ref{table:distance} shows the benefit of \methodName over all forms of distance functions $\Delta$ while MSE distance outperforms others reaching $82.4\%$ on classification of CIFAR-100 with ResNet18 backbone.
\begin{table}[t]
\caption{\textbf{Ablative Study on Different Distance Functions.} On ResNet-18 CIFAR-100 experiments, all variants of \methodName are better than baseline while \methodName with MSE as distance function gains the most. } 
\label{table:distance}
\centering
\begin{tabular}{c|c|c|c|c}
\toprule
\bf{\quad Baseline\quad\quad}& 
\bf{KL-Divergence}&
 \bf{\quad InfoNCE\quad\quad}&
 \bf{\quad\quad L1 \quad\quad\quad}&
 \bf{\quad MSE\quad\quad}\\ \hline\hline 
 79.6 & 81.5 & 80.5 & 80.5 & \bf{82.4}\\
 \bottomrule
\end{tabular}
\end{table}

\section{Conclusion}
We presented Deep Companion Learning (DCL), a novel DNN training approach that not only penalizes inconsistencies in model predictions but also leverages historical data to enhance generalization. Our strategy, centered around a deep-companion model (DCM), makes use of past predictions to enforce consistency and infer a meaningful semantic structure from the data. This novel mechanism introduces a dynamic, data-dependent regularization, optimizing both model consistency and the quality of representation in the logit space for improved class separability. Intuitively, DCL improves generalization during training by inferring a semantic structure for each class, and presenting the consistently reinforcing the confusing cases to the model. Our contributions include an Efficient Consistency Predictor that utilizes the companion model to dynamically adapt to new data, effectively minimizing prediction deviations. The empirical validation of DCL across various datasets and architectures demonstrates its capability to achieve state-of-the-art performance. Notably, it achieves comparable accuracy to models trained with pre-training while reducing computational demands. This highlights DCL's efficiency in training deep learning models from scratch. Furthermore, DCL is complementary to other settings including masked auto-encoders, fine-tuning, semi-supervised learning as well other methods using different augmentations.

\section*{Acknowledgments}
We acknowledge the support by the Air Force Research Laboratory grant FA8650-22-C1039, Army Research Office grant W911NF2110246, and the National Science Foundation grants CCF-2007350 and CCF-1955981 and CPS Medium 2317079.

%
%
\bibliographystyle{splncs04}
\bibliography{main}

\newpage
\section{Appendix}
In this supplementary document, we provide additional implementation details and additional ablative experiments. 
\subsection{Implementation Details}
\subsubsection{Dataset Details} \label{appendix.dataset_details}
Below, we describe all the datasets used in the main text along with the exact data augmentations. We use the same processing for all our experiments. 

\noindent\textbf{CIFAR-100 \cite{Krizhevsky09learningmultiple}} contains $50$K training and $10$K test images from $100$ classes with size $32\times32\times3$. We use basic data augmentations: `RandomCrop`, `RandomHorizontalFlip`, and `Mean-Std-Normalization`. 

\noindent\textbf{Tiny-ImageNet \cite{le2015tinyImagenet}} consists of  $100$K training and $10$K test images from $200$ classes with size $64\times64\times3$. Note that it is a subset of the popular ImageNet-1K dataset \cite{russakovsky2015imagenet}. We use the same augmentations as the CIFAR-100 dataset.

\noindent\textbf{ImageNet-1K \cite{russakovsky2015imagenet}} contains around $1.2$M training and $100$K test images from $1000$ classes with size $224\times224\times3$. We use `RandomCrop`, `RandomHorizontalFlip`, and `Mean-Std-Normalization`. In addition, we include the `RandAugment` implementation from the timm repository \cite{rw2019timm}.

\noindent\textbf{Fine-tunning Datasets.} We directly use dataset class from PyTorch of datasets CUB \cite{CUB}, Oxford-Pet \cite{Pets}, Food-101 \cite{Food-101} and Stanford Car \cite{cars} for fine-tunning tasks. We only apply base augmentations same as CIFAR-100 dataset. Same image sizes are applied for the fine-tunning datasets with the pre-trained dataset.

\subsubsection{Model Details on Different Problem Settings}
\subsubsection{Supervised Learning.}For supervised classification from scratch, we evaluate ResNet18 \cite{ResnetOriginalPaper},  ResNet34, ResNet50 and ShuffleNetV2 \cite{Ma_2018_ECCV_ShuffleNetV2} on CIFAR-100 experiments. We evaluate ResNet18, ResNet50, ViT-t\cite{dosovitskiy2020image} and Swin-t \cite{liu2021swin} architectures on ImageNet-1K experiments. The transformer-based architectures implementations are from the timm repository \cite{rw2019timm}.

\subsubsection{Fine-tunning.}For fine-tunning experiments, we use PyTorch provided ImageNet-1K pre-trained EfficientNet-v0 as initial model.

\subsubsection{Semi-Supervised Learning.} We follow FixMatch\cite{FixMatchConsistency} using WRN-28-2 backbone with default hyperparameters.

\subsubsection{Self-Supervised Pre-training.} For MAE\cite{he2022masked} pre-training experiments, due to computational expense, we only conduct experiments on ViT-B backbone. We will make the pre-trained backbone of \methodName open-source to facilitate potential downstream tasks.

\subsubsection{Knowledge Distillations.} We follow standard experiments in knowledge distillation literatures\cite{chen2022simkd,chen2021reviewkd,zhao2022decoupled}. We use ResNet8$\times$4 as student and trained ResNet32$\times$4 teacher model provided by official code of \cite{chen2022simkd}. 

\subsubsection{Optimizer}
\methodName is a novel algorithm for training a neural network which can be applied to any optimizers. We follow \cite{zhang2023gradient} using SGD optimizer for all ResNet experiments and AdamW on all transformer-based experiments. \methodName can improve performance and is compatible with any optimizers.

\subsubsection{Hyper-parameters} \label{appendix.hyper_parameter_details}
The unique hyper-parameter we need to tune is $\alpha$. We show an ablation study on $\alpha$ in Figure 6 in the main paper. We use $\alpha=0.6$ across all experiments. For experiments of CIFAR-100 and TinyImageNet, We follow \cite{zhang2023gradient} to train these models up to 200 epochs with cosine learning rate decay with 0.1 as the initial learning rate and batch size of 128. We run the ImageNet-1K experiments for 90 epochs for ResNet experiments and 300 epochs with transformer baselines. We follow \cite{zhang2023gradient} using batch size 256 for all ImageNet experiments due to limited computational resources. Some of the baselines\cite{du2022sharpness} reports slightly better performance using a much larger batch size $4096$. For fair comparison, we report the performance of reruns their official codes with batch size 256 with a careful tunning of hyperparameters.

We set $\eta_\omega=\eta_\theta$ over all experiments since $\eta_\omega$ and $\alpha$ both adjust the pace of the companion model to keep up with the deployed model. Under the best configuration $\alpha =0.6$ on ResNet18 and CIFAR100, we tried out $\eta_\omega=2 \eta_\theta$ and $\eta_\omega=\frac{1}{2}\eta_\theta$, reaching $81.4\%$ and $81.6\%$ not exceeding $82.4\%$ with $\eta_\omega=\eta_\theta$.

\subsection{Additional Experiments}
\subsubsection{Different Augmentations}
\methodName is a novel method for training a deep neural network that is effective regardless of data augmentation techniques. Good data augmentation can improve \methodName, serving as a complementary technique. TrivialAugment(TA)\cite{muller2021trivialaugment} is one of the SOTA data augmentation techniques. Tab.\ref{tab:aug} shows DCL uniformly gets better accuracy across different augmentations. Similar to other ImageNet experiments, we follow GAM[43] using batch size 256 due to limited computational resources. 
\begin{table}[htb]
    \caption{\textbf{Performance Comparison with Different Agumentations}}
    \label{tab:aug}
    \begin{subtable}[t]{.5\textwidth}
    \tablestyle{0.8pt}{1}
    \setlength{\tabcolsep}{10pt}
        \caption{WRN28-10 Accuracy on CIFAR-100}
        \centering
    \begin{tabular}{l|c c}
    \toprule
    \bf{ CIFAR100}&  
    \bf{CE}&
     \bf{DCL}\\
     \hline
    Basic &  81.4 & \bf 84.2\\
    \hline
     TA  & 84.3 & \bf 85.7\\
     \bottomrule
    \end{tabular}
    \end{subtable}%
   \begin{subtable}[t]{.5\textwidth}
   \tablestyle{0.8pt}{1}
    \setlength{\tabcolsep}{10pt}
        \centering
        \caption{ResNet50 Accuracy on ImageNet}
    \begin{tabular}{l|c c}
    \toprule
    \bf{ImageNet}&  
    \bf{CE}&
     \bf{DCL}\\
     \hline
         Basic &  76.0 & \bf 77.1\\
    \hline
     TA  & 76.2 & \bf 77.3\\
     \bottomrule
    \end{tabular}
    \end{subtable}
\end{table}

\subsubsection{Reduction of Computational Cost}
\subsubsection{Deep Companion Prototype}
Although an additional copy of the model is While previous works such as DML[44] and PS-KD[22] often employ an additional copy of the model, this approach indeed incurs extra computational costs. To reduce the computational cost, we proposed another variant of DCL namely, Deep Companion Prototype (DCP). DCP follows the same idea of DCL penalizing inconsistencies in historical model predictions. Instead of employing gradient methods to update the additional companion model (Line 7 in Algorithm 1), DCP maintained updated prototype logits for each class. For each class $k$ among the $K$ classes, we define a prototype $\bp^k\in \mathbb{R}^k$ in logits space. $\bp^k\in \mathbb{R}^k$ is dynamically updated along training. Then the regluarizer is penalizing the distance between current logits and its corresponding the prototype. Equation \ref{eq:update_theta_proto} shows the update of the deployed model.
\begin{align}
\label{eq:update_theta_proto}
    \btheta_{t+1}= \btheta_t - \eta_{\btheta} \nabla_{\btheta} (\mathcal{L}(\btheta) + \frac{1}{|B_t^k|}\sum_{k\in[K]}\sum_{(\bx,y)\in B_t^k}\Delta (f(\btheta,\bx), \bp^k_t))
\end{align}
where $k$ $(\bx,y)\in B_t^k$ denotes the data for class $k$ in current batch $B_t$ at iteration $t$. Equation \ref{eq:proto} shows the update of the prototype combines combines historical logtis prototype and the current mean of logits of class $k$. If current batch does not contain class $k$ data, we then do not update the prototpye.
\begin{align}
\label{eq:proto}
    \bp^k_{t+1} = \begin{cases}
    \alpha \bp^k_{t} +   (1-\alpha) \frac{1}{|B_t^k|}\sum_{(\bx,y)\in B_t^k}f(\btheta, \bx),&\quad |B_t^k|\neq 0  \\
    \bp^k_{t}, &\quad |B_t^k|=0\\
\end{cases}
\end{align}

Table \ref{table:dcp} demonstrates DCP outperforms recent baselines, but falls short of DCL. While it do not require updates of additional companion model,  thereby reducing computational costs compared to DCL.

\begin{table}[t]
\tablestyle{8pt}{1.2}
\caption{\textbf{Comparison of DCP with Recent Baselines and DCL on CIFAR100.} DCP outperforms most recent baselines but falls short of DCL performance. } 
\label{table:dcp}
\centering

\begin{tabular}{l|c|c}
\toprule

\textbf{Methods} & \textbf{\begin{tabular}[c]{@{}c@{}}  ResNet18  \end{tabular}}   & \textbf{\begin{tabular}[c]{@{}c@{}} 
  WRN28-10   \end{tabular}} 
 \\ \hline\hline 
 CE\cite{zhang2023gradient}     & $78.3\pm 0.32$ & $81.4\pm 0.13$ \\ 
 DML  \cite{zhang2018deep}    & $79.9\pm 0.32$  & $82.7\pm0.14$ \\ 
 SAM  \cite{du2022sharpness}  & $79.3\pm 0.25$  & $83.4\pm0.06$ \\ 
 PSKD \cite{kim2021self}   & $80.6\pm 0.26$  & $81.9\pm0.10$  \\ 
SAF  \cite{du2022sharpness}  & $80.8\pm 0.08$  & $83.8\pm0.04$ \\ 
 GAM  \cite{zhang2023gradient}    & $80.5\pm 0.24$  & $83.5\pm0.09$ \\ 
 \hline
  Ours (DCP)   & $\mathbf{81.0\pm 0.21}$   & $\mathbf{83.8\pm0.09}$ \\ 
 Ours (DCL)   & $\mathbf{82.4\pm 0.26}$   & $\mathbf{84.2\pm0.10}$ \\ 
\bottomrule
\end{tabular}
\end{table}
\begin{small}
\begin{table}[h]
\tablestyle{1pt}{1}

\caption{\textbf{DCL with smaller companion models.} On ResNet-18 CIFAR-100 experiments, we apply different-sized companion models. Even the companion model is very small, DCL can still improve over CE baselines.} 
\label{table:small}
\centering

\begin{tabular}{l|c|c|c|c|c|c}
\toprule
$\bomega$ Models &{ResNet18}& 
{ResNet10}&
{ResNet10-l}&
{ResNet10-m}&
{ResNet10-s} &
{ResNet10-xs}\\ \hline\hline 
$\btheta$ Accs & 82.4 & 81.4 & 80.8 & 80.4 & 80.4 & 80.2\\
$\bomega$ Accs & 80.4 & 76.9 & 68.9 & 53.3 & 33.3 & 23.7\\
 \bottomrule
\end{tabular}
\end{table}
\end{small}

\subsubsection{DCL with smaller companion models.}
The companion network in DCL mirrors the architecture of the primary deployed model. Since the interaction between the deployed model and companion model only happens in the logits space, we can use a companion model with smaller architectures. Table \ref{table:small} shows experiments with ResNet-18 CIFAR 100 experiments with smaller models. Even the companion model is very small, DCL can still improve over CE baselines.

\begin{table}[h]
\tablestyle{8pt}{1.2}
    \caption{\textbf{Train Companion Model with Less Data.}On ResNet-18 CIFAR-100 experiments, we apply different-sized subbatch to companion models. Even we use only $10\%$ of data to train the companion model, DCL can still improve over CE baselines.}
    \label{tab:partial}
    \centering
    \begin{tabular}{l| c | c | c | c | c}
    \toprule
    \bf{\% of data}&  
    {10\%} & 20\% & 50\% & 80\% & 100\%\\
     \hline
    Acc(\%)& 80.7 & 80.9  & 81.4 & 81.8 & 82.4\\
     \bottomrule
    \end{tabular}
\end{table}

\subsubsection{DCL with less data for training the companion model.}
The companion network can be trained by a subset of training data. At each iteration, we can sample a portion of the batched data and train the companion model. Table \ref{tab:partial} shows experiments with ResNet-18 CIFAR 100 experiments fewer data for the companion training. It shows DCL with less data to train the companion model which saves computations but is still beneficial.

\begin{table}[t]
\tablestyle{0.8pt}{1}
\setlength{\tabcolsep}{4pt}
\caption{\textbf{Different $\alpha$ Values on CIFAR-100 ResNet18 Experiments.} As $\alpha$ approaches 1, the companion model updates slowly, capturing more information of early under-trained model, resulting in suboptimal performance. When $\alpha$ is close to 0, the companion model fails to effectively capture historical model information and reduce to the current instantaneous model.}
\label{tab:alpha}
\centering
\begin{tabular}{l|c|c|c|c|c|c|c|c|c|c|c|c}
\toprule
\textbf{$\alpha$} value&0& 0.01 &0.1& 0.2 &0.3 &0.4 &0.5 &0.6 &0.7 &0.8 &0.9 & 0.99\\
\hline\hline
 $\btheta$ Acc ($\%$)& 80.9 & 81.2 & 81.3 &81.6 & 81.7 & 81.8 &81.8 & \bf{82.4} & 82.0 &81.1& 80.8 & 80.0 \\
 $\bomega$ Acc ($\%$)& 80.6 & 80.4 & 80.3 & 80.7 & 80.4 & 80.6 & 80.9 & 80.4 & 80.0 & 76.9 &63.94 & 1.0\\
 \bottomrule
\end{tabular}
\end{table}

\subsubsection{Ablative Study on  $\alpha$ }
\noindent \textbf{Different $\alpha$ Values.} Table \ref{tab:alpha} illustrates the performance of \methodName for different $\alpha$ values. As $\alpha$ approaches 1, the companion model updates slowly, capturing information of early undertrained model, resulting in suboptimal performance. When $\alpha$ is close to 0, the companion model fails to effectively capture historical model information and reduces to the current instantaneous model. Thus a proper choice in between maximizes the benefits of \methodName. 

\noindent \textbf{Fixed $\alpha$ vs. $\alpha_t$}
 We use a fixed $\alpha$ for implementation to avoid weighing much of the noisy output of the early stage. We ran DCL with $\alpha_t=\frac{t-1}{t}$, getting worse results, e.g. $81.5\%$ on CIFAR100 with ResNet18.

\subsubsection{Comparison with esembling methods}
\begin{table}[t]
\tablestyle{8pt}{1.2}
\setlength{\tabcolsep}{1pt}.
    \caption{\textbf{Performance Comparison with Baselines of Esembling Methods with ResNet-18 on CIFAR-100.}}
    \label{tab:esb}
    \centering
    \begin{tabular}{l| c c c |c}
    \toprule
    \bf{N}&  
    {Voting} & Bagging & Snapshot &  DCL\\
     \hline
    N=2&78.9 & 78.4 & 78.3 &\multirow{ 2}{*}{\bf 82.4} \\
    N=10& 81.0 & 79.3 &78.9&\\
     \bottomrule
    \end{tabular}
\end{table}

Since DCL is a method that requires training two models, it is reasonable to compare it with a resembling methods baseline. Table \ref{tab:esb} shows a comparison of DCL with esembling baselines, including basic voting, bagging and snapshot esembling\cite{huang2017snapshot}. DCL is better than all ensembling methods even if we set the number of estimators as large as 10. Compared to ensembling, our approach requires fewer training computations to achieve better performance and allows more efficient inference by directly testing on the deployed model.

\end{document}